\documentclass[letterpaper, 10 pt, conference]{ieeeconf} 

\IEEEoverridecommandlockouts    

\usepackage{graphics} 
\usepackage{epsfig} 
\usepackage{array}
\usepackage{float} 
\usepackage[T1]{fontenc}
\usepackage{xcolor,soul}
\usepackage{bm}
\usepackage{subfigure}
\usepackage{amsfonts,amssymb, amsmath}
\usepackage{cases}
\usepackage{graphicx}
\usepackage{booktabs}
\usepackage{multirow}
\usepackage{multicol}
\usepackage{hyperref}
\hypersetup{hidelinks,
	colorlinks=true,
	allcolors=black,
	pdfstartview=Fit,
	breaklinks=true}
\usepackage{mathtools}
\newcommand{\MYhref}[3][blue]{\href{#2}{\color{#1}{#3}}}%
\newcounter{RNum}
\renewcommand{\theRNum}{\arabic{RNum}}
\newcommand{\Remark}{\noindent\textit{\textbf{Remark}~\refstepcounter{RNum}\textbf{\theRNum}: }}
\soulregister\Remark7
\title{\LARGE \bf
RoboDexVLM: Visual Language Model-Enabled Task Planning and Motion Control for Dexterous Robot Manipulation
}
\author{Haichao Liu, Sikai Guo, Pengfei Mai, Jiahang Cao, Haoang Li, and Jun Ma, \textit{Senior Member, IEEE}
\thanks{Haichao Liu, Sikai Guo, Pengfei Mai, Jiahang Cao, Haoang Li, and Jun Ma are with The Hong Kong University of Science and Technology (Guangzhou), China (e-mail: jun.ma@ust.hk).}
}

\begin{document}

\maketitle
\begin{abstract}
This paper introduces RoboDexVLM, an innovative framework for robot task planning and grasp detection tailored for a collaborative manipulator equipped with a dexterous hand. Previous methods focus on simplified and limited manipulation tasks, 
which often neglect the complexities associated with grasping a diverse array of objects in a long-horizon manner. 
In contrast, our proposed framework utilizes a dexterous hand capable of grasping objects of varying shapes and sizes while executing tasks based on natural language commands. The proposed approach has the following core components: First, a robust task planner with a task-level recovery mechanism that leverages vision-language models (VLMs) is designed, which enables the system to interpret and execute open-vocabulary commands for long sequence tasks. Second, a language-guided dexterous grasp perception algorithm is presented based on robot kinematics and formal methods, tailored for zero-shot dexterous manipulation with diverse objects and commands.
Comprehensive experimental results validate the effectiveness, adaptability, and robustness of RoboDexVLM in handling long-horizon scenarios and performing dexterous grasping. These results highlight the framework's ability to operate in complex environments, showcasing its potential for open-vocabulary dexterous manipulation. Our open-source project page can be found at \MYhref[blue]{https://henryhcliu.github.io/robodexvlm}{https://henryhcliu.github.io/robodexvlm}.
\end{abstract}

\section{Introduction}
Robotic manipulation has become a cornerstone of modern technological progress, driving advancements in manufacturing, healthcare, and domestic automation. By bridging perception, reasoning, and physical interaction, these systems enhance productivity, enable safe operation in hazardous environments, and address critical societal challenges such as labor shortages.
A robotic manipulation pipeline typically involves four core stages: environment perception via sensors (e.g., LiDAR, RGB-D cameras), object detection to localize targets, grasp perception to compute stable contact points, and motion planning to execute collision-free trajectories. For complex, long-horizon tasks, such as organizing cluttered spaces or preparing meals, the system must decompose abstract goals into actionable sequences, which require a semantic understanding of object affordances and task dependencies.

Recent advances in visual perception have enabled language-guided object detection and segmentation, which are critical for robotic manipulation. While traditional detectors like YOLOv11~\cite{make5040083} excel in speed, their reliance on labeled datasets limits adaptability. Modern frameworks like Grounding DINO~\cite{liu2024grounding} and SAM~\cite{kirillov2023segment} address this via zero-shot generalization: SAM achieves prompt-agnostic segmentation, while Grounding DINO bridges text prompts (e.g., ``blue cube'') to object localization. Hybrid approaches like LangSam~\cite{luca-medeiros_lang-segment-anything}, integrating these models, further enable real-time mask generation without human refinement, which is paramount for dynamic manipulation tasks. Building upon such perceptual foundations, recent works for robot manipulation focus on translating segmented objects into executable action sequences.
\begin{figure*}[t]
    \centering
    \includegraphics[width=0.85\linewidth]{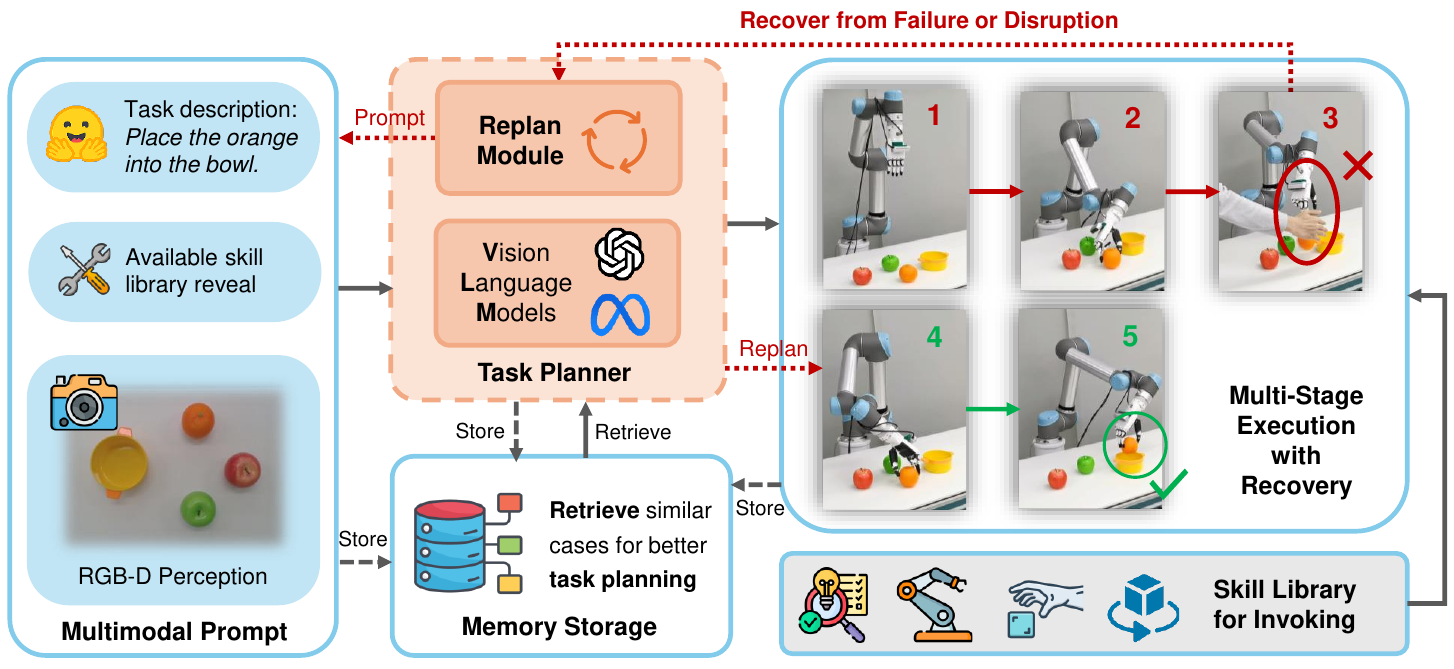}
    \caption{Overview of our RoboDexVLM. The multimodal prompt, comprising human command, available skill list, RGB-D image, and relevant memory items, is transmitted to the VLM for task planning. Upon receiving the skill invoking sequence from the VLM, the dexterous robot executes the skills until task completion. Dashed lines indicate the recovery process following failed operations.}
    \label{fig:operationDemo}
\end{figure*}

In terms of task planning for robot manipulation, the integration of large language models (LLMs) into task planning has redefined robotic manipulation, enabling systems to interpret abstract instructions and translate them into actionable sequences~\cite{jin2024robotgpt}. 
Recent works leverage vision-language models (VLMs) to represent manipulation tasks as constraint satisfaction problems~\cite{gao2024physically,firoozi2023foundation}. For instance, ReKep~\cite{huang2024rekep} employs VLMs to map 3D environmental keypoints to a numerical cost function, optimizing grasp poses by minimizing collision risks and instability. This approach bypasses traditional heuristic-based planners, enabling data-efficient adaptation to novel objects. Similarly, OmniManip~\cite{pan2025omnimanip} introduces an object-centric interaction representation that aligns VLM-derived semantic reasoning with geometric manipulation primitives. By constructing a dual closed-loop system for planning and execution, OmniManip achieves zero-shot generalization across diverse tasks using a parallel gripper without requiring VLM fine-tuning. 
Further advancing this paradigm, RoboMamba~\cite{liu2024robomamba} utilizes a vision-language-action (VLA) model to predict target object poses and transformation matrices directly from language prompts.
Complementary approaches, such as CLIPort~\cite{shridhar2022cliport} and VoxPoser~\cite{pmlr-v229-huang23b}, demonstrate how VLMs can synthesize spatial affordance maps or generate code-like action scripts, respectively, to guide manipulators in complex tasks like assembly or kitchen operations.
These innovations collectively highlight a shift toward language-grounded manipulation, where semantic understanding and geometric reasoning are tightly coupled. By unifying perception, planning, and execution under a VLM-driven framework, robotic systems gain the flexibility to interpret ambiguous commands for real-world deployment~\cite{ha2023scaling}.

However, the aforementioned robot manipulation methods merely use parallel grippers to simplify the challenging operation problem with a high degree of freedom (DoF). In contrast, dexterous hands, characterized by their multi-fingered design and human-like articulation~\cite{liu2008multisensory,kim2021integrated}, are emerging as a means for robots to adaptively execute grasping operations in real-world scenarios.
Unlike parallel-jaw grippers, which excel in rigid, predefined grasps but struggle with delicate or deformable objects~\cite{liu2021research}, dexterous hands emulate the adaptability of human manipulation~\cite{dogar2010push}. They enable in-hand reorientation and contact-rich interactions, making them indispensable for tasks such as tool use or utensil handling in cluttered environments.
Recent advancements in grasp perception have largely focused on parallel grippers, yielding models like GG-CNN~\cite{morrison2020learning}, Contact-GraspNet~\cite{sundermeyer2021contact}, and AnyGrasp~\cite{fang2023anygrasp}, which predict antipodal grasps from point clouds or RGB-D data. 
By contrast, native dexterous grasp perception demands precise prior geometric information. Approaches like $\mathcal{D}(\mathcal{R}, \mathcal{O})$ grasp~\cite{wei2024mathcaldro} unify robot-object interaction representations across embodiments, enabling cross-platform grasp synthesis but relying heavily on accurate object meshes and joint torque constraints. To circumvent the challenges of model-based planning, many researchers adopt reinforcement learning (RL) and imitation learning trained on motion-capture data from human demonstrations~\cite{chen2022learning,han2024learning}. Works such as DexCap~\cite{wang2024dexcap} introduce scalable systems for collecting high-fidelity hand-object interaction data with RL policies to achieve goal-conditioned dexterous grasping.
Despite these strides, a critical gap remains: current methods predominantly focus on grasp pose generation, neglecting the integration of dexterous manipulation with task-level planning for embodied AI~\cite{chen2023bi,ma2024survey}.  
Bridging dexterous manipulation with the VLM-driven task planners represents an untapped frontier, where zero-shot generalization and contextual adaptability could unlock unprecedented versatility in robotic systems.

Lastly, ensuring robust recovery from failures is essential for deploying robotic manipulation systems in real-world settings, where perceptual ambiguities, environmental uncertainties, and imperfect model outputs frequently lead to errors~\cite{fang2023anygrasp,ak2023learning}. Specifically, AIC MLLM~\cite{xiong2024aic} integrates test-time adaptation, allowing agents to dynamically adjust their perception and planning modules based on real-time analysis in the selected scenarios. 
Therefore, self-assessment mechanisms hold the potential to enhance resilience in unstructured environments~\cite{pan2025omnimanip}, highlighting the significance of closed-loop adaptability for reliable, long-term robotic operations. Nonetheless, the recovery from failure through task re-planning in robotic manipulation remains an area of ongoing exploration.

To address the above research gaps, we propose RoboDexVLM, an open-vocabulary task planning and motion control framework for dexterous manipulation, as illustrated in Fig.~\ref{fig:operationDemo}. 
By integrating visual data with natural language instructions, VLMs enable robots to interpret intent, infer task hierarchies based on the given skill library, and adapt plans dynamically. When paired with dexterous hands capable of versatile object interaction, this synergy unlocks precise, context-aware manipulation, paving the way for robots to operate seamlessly in unstructured, real-world environments. The contributions are summarized as follows:

\begin{itemize}
    \item We propose RoboDexVLM, a novel framework that integrates a VLM-based automated task planning pipeline with a modular skill library to achieve long-horizon dexterous manipulation, which effectively bridges high-level planning and low-level kinematic constraints.

    \item The framework integrates VLMs to enable primitive-based task decomposition and execution from open-vocabulary commands. A robust task planner dynamically interprets user intent, optimizes grasp poses, and incorporates failure recovery mechanisms leveraging its reflection ability for long-horizon adaptability.

    \item Through extensive real-world experiments, we demonstrate its effectiveness in complex environments, highlighting its stability in dexterous grasping, adaptability to novel objects, and resilience to unfamiliar tasks.
\end{itemize}

\section{Language-Grounded Manipulation with Canonical Primitives}
\label{subsec:interaction_primitives}
\subsection{Task Planning with Manipulation Primitives}
The RoboDexVLM framework represents a significant advancement in the field of robotic manipulation by seamlessly bridging the gap between high-level task planning and low-level execution through an innovative structured skill library. This library is at the heart of enabling zero-shot manipulation capabilities, where robots can perform tasks they have not been explicitly programmed for, solely based on natural language instructions $\mathcal{L}$, e.g., \textit{open the box and place the bigger carambola inside} as shown in Fig.~\ref{fig:operationDemo}. The architecture of this skill library $\mathcal{S}=\{\mathcal{F}_1(X_1), \mathcal{F}_2(X_2), \cdots, \mathcal{F}_n(X_n)\}$ is meticulously designed to formalize manipulation primitives while maintaining enough flexibility to allow for adaptation guided by language inputs. In $\mathcal{S}$, each skill unit $\mathcal{F}_i$ has its required input $I_i$ to activate specific actions.

\begin{figure}[t]
    \centering
    \includegraphics[width=1\linewidth]{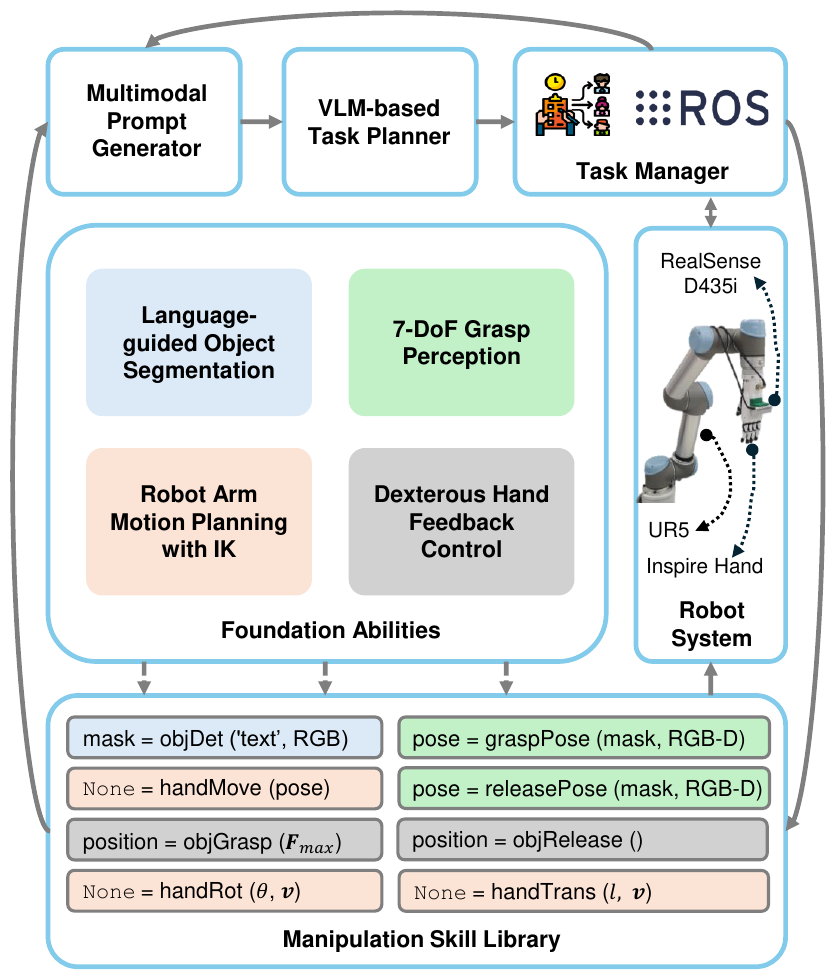}
    \caption{Working pipeline of the proposed RoboDexVLM. The system comprises several complementary modules designed to facilitate a closed-loop manipulation framework. The task manager orchestrates the execution of $\mathcal{F}_i$ based on $\mathcal{O}$ generated by the VLM. Skills are performed through the grounding of the four foundational capabilities established at the core of the system.}
    \label{fig:pipeline}
\end{figure}

As illustrated in Fig. \ref{fig:pipeline}, the skill library $\mathcal{S}$ comprises eight atomic skills that encapsulate fundamental manipulation actions such as detecting, grasping, moving, and placing objects. These atomic skills $\mathcal{F}_i,i\in\{1,2,\cdots,8\}$ are the building blocks upon which complex manipulation tasks commanded by $\mathcal{L}$ can be constructed. Each skill is designed to operate independently yet cohesively within the overall framework, ensuring smooth transitions and efficient execution of tasks. 

Leveraging the world knowledge and reasoning ability, a VLM is utilized to generate the skill order and the corresponding required function inputs by the following prompt-reasoning process: \begin{equation}\label{eq:llmInfoFlow}
    \left\{\mathcal{R}_\tau,\mathcal{O}_\tau,\mathcal{I}_\tau\right\} = \mathcal{T}\left(K\left(\mathcal{S},\mathcal{M}_\tau, \mathcal{L}_\tau\right)\right),
\end{equation}
where the input elements of the context generator $K(\cdot)$ are constant system message $\mathcal{S}$ with Chain-of-Thought (CoT) reasoning template, memory message $\mathcal{M}_\tau$, and the human message $\mathcal{L}_\tau$ as the task description at time step $\tau$. We denote the function $\mathcal{T}$ as the reasoning process of the VLM. The output of the VLM agent is designed as three folds: The CoT reasoning text $\mathcal{R_\tau}$, the primitive order of the skills to be invoked $\mathcal{O}_\tau=\{O_1,O_2,\cdots,O_m\}, O_i\in\mathcal{S}$, and the corresponding input primitive of the skills \( \mathcal{I} = \{ I_1, I_2, \ldots, I_m \} \), where \( I_i \in \text{set}(X_i) \) for \( i \in \{1, 2, \cdots, 8 \}\). Note that $\mathcal{R}_\tau$ enhances the transparency of the designed skill order by the VLM, and it is profoundly beneficial for memory reflection for few-shot learning. 

\subsection{Interaction Primitives for Skill Execution}
A key aspect of the RoboDexVLM framework's design is its standardized input-output interfaces $\mathcal{F}_i(X_i)$ for each skill. These interfaces facilitate seamless interaction between the different phases, allowing the VLM to dynamically chain them together based on the specific requirements of the language command $\mathcal{L}_\tau$. Therefore, we maintain dynamic variable storage formalized as
\begin{equation}
    \mathcal{D} = \left\{\mathcal{E}_\text{lang},\mathcal{P}_\text{RGB},\mathcal{P}_\text{Depth},\mathcal{B}_\text{img},\mathcal{G},F_\text{max}, A\right\},
\end{equation}
where $\mathcal{E}_\text{lang}$ denotes the language guidance for image segmentation, pixel matrices of RGB-D images are expressed as $\mathcal{P}_\text{RGB}\in \mathbb{R}^{H\times W\times 3}$ and $\mathcal{P}_\text{Depth}\in \mathbb{R}^{H\times W\times 1}$, respectively. The binary result of semantic segmentation for $\mathcal{E}_\text{lang}$ is stored in $\mathcal{B}_\text{image}\in \mathbb{B}^{H\times W\times 1}$. For object grasping, $\mathcal{G}$ contains necessary geometric values, and it is elaborated in Section~\ref{subsec:dexterousManipulationTransfer}. The maximum contact force with objects is indicated by $F_\text{max}$, while the geometric vector for robot motion (e.g., rotate and twist) is denoted as $A=\{d,\theta,r\}\in \mathbb{R}^{3}$.
Furthermore, the skill functions $\mathcal{F}_i$ illustrated in the lower part of Fig.~\ref{fig:pipeline} can query the variable storage $\mathcal{D}$ to retrieve updated data for real-time manipulation.

\section{Skill Execution with Dexterous Manipulation}
\label{subsec:skill_execution}
Combining with the primitive order of the skills $\mathcal{O}_\tau$, the robot system executes dexterous manipulation in sequence and simultaneously updates the variable storage $\mathcal{D}$ until the long-horizon task from the initial language instruction $\mathcal{L}$ is completed.

\subsection{Perception-Action Paradigm}
\label{subsec:pipeline}

The perception-action paradigm of the RoboDexVLM framework is designed to achieve precise and robust manipulation through a closed-loop execution system, as illustrated in Fig. \ref{fig:pipeline}. This pipeline integrates several advanced technologies to achieve the foundation abilities supporting the skill invoking for diverse tasks.
In each foundation ability, the RoboDexVLM system follows the coherent perception-action paradigm. 

First, the language-guided image segmentation module generates semantic-level object masks by integrating linguistic embeddings $\mathcal{E}_\text{lang}$ with real-time visual input $\mathcal{P}_\text{RGB}$. Inspired by~\cite{luca-medeiros_lang-segment-anything}, we employ two complementary models for semantic mask generation.
Initially, an open-set object detector, Grounding DINO~\cite{liu2024grounding}, performs zero-shot text-to-bounding-box detection by aligning $\mathcal{E}_\text{lang}$ (e.g., \textit{the red apple}) with visual features extracted from $\mathcal{P}_\text{RGB}$. The region proposal score for a bounding box $\bm B_i$ is computed, and the ones exceeding a threshold $\tau_d$ are retained as $\{\bm B^* = {\bm B_i \,|\, \forall i,\, \mathrm{\small Score}(\bm B_i) > \tau_d}\}$.
Subsequently, SAM~\cite{ravi2025sam} refines $\bm B^*$ into pixel-precise masks ${\mathcal{B}_\text{img}}$, ensuring instance-aware segmentation even in cluttered or strange scenes.
This hybrid approach enhances both efficiency (reducing search space via coarse-to-fine processing) and scene comprehension robustness under ambiguous instructions.

Second, the system leverages $\mathcal{P}_\text{RGB}$ and $\mathcal{P}_\text{Depth}$ aligned with segmented masks ${\mathcal{B}_\text{img}}$ to filter target objects, followed by inferring optimal grasping pose for the end-effector of the robot via AnyGrasp~\cite{fang2023anygrasp}. For each candidate grasp pose hypothesis $\mathcal G_j$, geometric-geometric alignment computes pairwise correspondence scores through cosine similarity:
\begin{equation}
    s(\mathcal G_p,\mathcal G_q) = \frac{\mathbf{f}_{\theta}(\mathcal G_p)^T , \mathbf{f}_{\theta}(\mathcal G_q)}{|\mathbf{f}_{\theta}(\mathcal G_p)|_2 ,  |\mathbf{f}_{\theta}(\mathcal G_q)|_2},
\end{equation}
where $\mathbf{f}_{\theta}(\cdot)$ encodes grasp poses into feature vectors via the learned network $\theta$. Aggregating similarities across hypotheses forms a correspondence matrix $\bm S_{pq}\in\mathbb{R}^{N\times N}$, where $N$ is the candidate number and raw confidence for candidate ${j}$ derives from row-wise summation: $\mathcal{C}_{j} = {\sum_{k=1}^N S_{jk}}.$ The optimal grasp $\mathcal{G}_{j}^*=\text{argmax}_j\,\mathcal{C}_{j}$ selects poses maximizing spatial consistency within local geometry constraints inferred from $\mathcal{P}_\text{Depth}$. Note that the pose for object placing and the final pose of an isolated action $A$ can be generated with a similar approach.

For action execution, the trajectories of the robot arm are calculated using Denavit-Hartenberg kinematics. Interpolated waypoints are optimized to maintain end-effector orientation constraints during the approach, ensuring smooth and stable movements throughout the task after each step of real-time perception. In a closed-loop manner, the robot can efficiently adjust its manipulation target for tasks in dynamic scenarios.
\subsection{Dexterous Manipulation Pose Generation}\label{subsec:dexterousManipulationTransfer}

Grasp perception methods for parallel grippers have significantly advanced in simplifying grasp synthesis, and these developments can be leveraged as foundational priors for dexterous manipulation.
Specifically, we transfer parallel-gripper grasp proposals to dexterous hands through kinematic retargeting.
This approach allows the utilization of existing perception frameworks while effectively accommodating the higher DoF in dexterous hands, leading to a higher success rate, especially when grasping objects with strange shapes (e.g., carambola). 

\begin{figure}[t]
    \centering
    \includegraphics[width=0.62\linewidth]{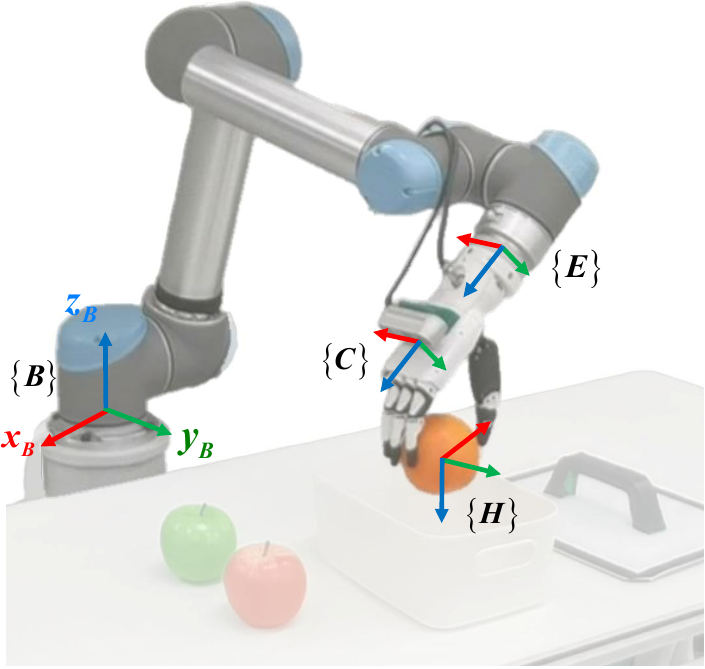}
    \caption{Robot manipulation system settings for RoboDexVLM. The robot manipulator (UR5) is supposed to grasp and manipulate the objects on the table and interact with the drawer or other kinds of containers using a dexterous hand (Inspire Hand) as the end effector with the perception generated by the RGB-D camera (RealSense D435i) mounted on the hand. The coordinates of the base $\{B\}$, hand $\{H\}$, end-effector $\{E\}$, and camera $\{C\}$ are illustrated accordingly.}
    \label{fig:experimentSettings}
\end{figure}

The grasp proposals are defined as  \( \mathcal{G} = \{\bm t, \bm R, w\} \), where $\bm t\in \mathbb{R}^{3}$ is the grasping center in Cartesian frame, $\bm R\in{SO(3)}$ is the rotation matrix, and $w$ is the width of the gripper for a successful grasping. 
The force-sensing module of the dexterous hand facilitates the simultaneous closure of all fingers once the desired pose is attained, and this closure continues until the applied force reaches the maximum threshold \( F_{\text{max}} \).

In order to utilize $\mathcal{G}$ in dexterous grasping, we need to determine the calibration matrix 
\begin{equation}\label{eq:EH-T}
    \prescript{{E}}{{H}}{\boldsymbol{T}}=
\prescript{{B}}{{E}}{\boldsymbol{T}}^{-1}
\prescript{{B}}{{H}}{\boldsymbol{T}}.
\end{equation}
from the flange frame $E$ to the corresponding dexterous hand frame $H$. First, the end pose of UR5 in the base frame $B$ can be calculated by
\begin{equation}\label{eq:BE-T}
    \prescript{{B}}{{E}}{\boldsymbol{T}}=
\prescript{{B}}{{C}}{\boldsymbol{T}}
\prescript{{C}}{{H}}{\boldsymbol{T}}
\prescript{{E}}{{H}}{\boldsymbol{T}}^{-1},
\end{equation}
where $\prescript{{B}}{{C}}{\boldsymbol{T}}$ and $\prescript{{E}}{{H}}{\boldsymbol{T}}$ are the eye-on-hand calibration matrix and hand-on-end calibration matrix, respectively. Furthermore, the grasping pose of the hand in the camera frame $C$ is expressed as
\begin{equation}
    \prescript{{C}}{{H}}{\boldsymbol{T}}= \begin{bmatrix} \bm{R} & \bm{t} \\ \bm{0}_{1 \times 3} & 1 \end{bmatrix}\in{SE(3)},
\end{equation}
with which $\prescript{{B}}{{E}}{\boldsymbol{T}}$ can be calculated by (\ref{eq:BE-T}). 

To get an accurate hand-on-end calibration matrix $\prescript{{E}}{{H}}{T}$, we move the pose of the dexterous hand manually and fine-tune it to ensure that the current poses of the thumb finger and middle finger can correspond to those of two-finger grippers shown in the 3D point cloud. Then the current $\prescript{{B}}{{H}}{T}$ is recorded, and finally, the calibration matrix can be calculated by (\ref{eq:EH-T}).
All the above-mentioned frames are illustrated in Fig.~\ref{fig:experimentSettings} to facilitate the understanding of the dexterous grasp pose generation process,

\subsection{Recovery Strategy from Failures During Manipulation}
\label{subsec:recovery}

To ensure robustness against execution errors, RoboDexVLM employs a dual-layer recovery mechanism. After each skill execution, success verification is conducted using depth-based change detection and position feedback from all the fingers of the dexterous hand. In case of failure, such as when grasp falls, the system constructs a reflection prompt
\begin{equation}
    \mathcal{H}_{\text{reflect}} = [E_{\text{error},\tau}, \mathcal{P}_{\text{RGB},\tau+1}, \mathcal{O}_{\text{history}}],
\end{equation}
containing details about the detected error $E_{\text{error}}$, the current scene state $\mathcal{P}_{\text{RGB}}$, and a history of previous skill attempts $\mathcal{O}_{\text{history}}\subset \mathcal{O}_\tau$.
This prompt is processed by the VLM using 
\begin{equation}\label{eq:recoverReflection}
    \left\{\mathcal{R}_{\tau+1},\mathcal{O}_{\tau+1},\mathcal{I}_{\tau+1}\right\} = \mathcal{T}\left(K\left(\mathcal{H}_\text{reflect}\right)\right),
\end{equation}
to propose adjusted skill sequences. For instance, if an object slips during a grasp attempt, the system might insert a \textit{HandRot} primitive to reorient the object for a more secure grip. To prevent infinite loops, the system resumes execution from the last successful skill and limits recovery attempts to three per task. Experimental results presented in Section \ref{sec:results} show that this strategy significantly increases the task success rate, especially in long-horizon manipulation scenarios.

\section{Experimental Analysis}
\label{sec:results}

\subsection{Environment Settings}

The experiments are conducted in a real-world environment designed for robotic manipulation tasks. 
As illustrated in Fig.~\ref{fig:experimentSettings}, this setup includes a UR5 robotic arm equipped with an Inspire 5-fingered dexterous hand, an Intel RealSense D435i RGB-D camera for object detection and scene analysis, and a workspace containing various objects of different shapes, sizes, and textures to test the versatility of the RoboDexVLM framework. We choose GPT-4o~\cite{hurst2024gpt} as our foundation model for dexterous robot operation. The system operates under varying desktop arrangements to ensure robust performance across different scenarios. All computations and model predictions run on a workstation with RTX 3080Ti GPU with 12\,GB graphical memory to support real-time processing requirements.

\subsection{Effectiveness of Zero-shot Dexterous Manipulation}
\begin{figure}[t]
    \centering
    \includegraphics[width=0.85\linewidth]{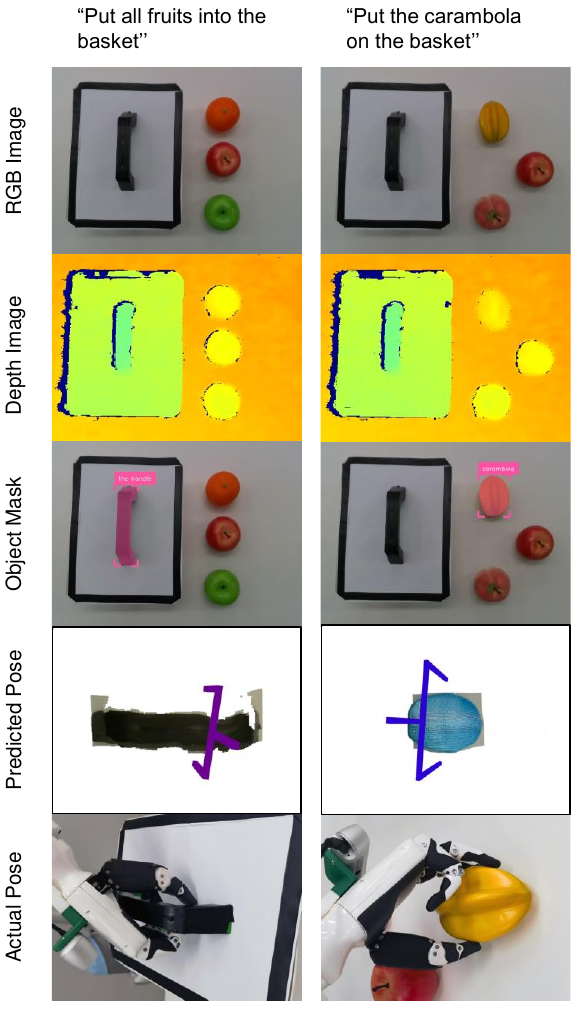}
    \caption{The dexterous grasping pose generation process. The object name for segmentation is provided at the top of the figure. In the RGB image with mask, the semantic segmentation masks of the object described by text are marked accordingly. The blue anchors in the images of the grasping pose area are the grasp perception results.}
    \label{fig:GraspingDemo}
\end{figure}

We qualitatively evaluate the dexterous manipulation performance of RoboDexVLM using the demonstration of intermediate processes from different open-vocabulary tasks. 

In terms of putting fruits into a lidded box, the robot is required to identify and grasp the handle of the lid accurately using natural language commands. As demonstrated in the first column of Fig.~\ref{fig:GraspingDemo}, the assigned object, the handle of the box lid, is masked correctly. With the above generated $\mathcal{B}_\text{img}$, from the grasp perception for paralleled grippers in the fourth row of Fig.~\ref{fig:GraspingDemo}, RoboDexVLM generates a normal grasp pose. Further, using the dexterous pose generation algorithm proposed in Section~\ref{subsec:dexterousManipulationTransfer}, the dexterous hand grasp the lid using a human-like grasp posture.

On the contrary, as illustrated in the second column of Fig.~\ref{fig:GraspingDemo}, when the open-vocabulary task is \textit{Put the carambola
on the basket}, the VLM generates the command of $\mathcal{E}_\text{lang}=\texttt{carambola}$ for the accurate object mask shown in the third row of Fig.~\ref{fig:GraspingDemo}. As shown in the fourth row of Fig.~\ref{fig:GraspingDemo}, the irregular shape of the carambola makes it susceptible to damage during grasping with parallel grippers, as the vertical contact between the two surfaces can easily compromise the fruit's integrity. In comparison, the transferred dexterous grasp pose for the hand establishes an enveloping surface, allowing the fruit to be held in a semi-encircled manner. Consequently, this grasping configuration not only mitigates excessive pressure from the vertical contact surfaces of parallel grippers but also enhances grasping stability.
\begin{figure*}[htbp]
\centering
\subfigure[Key frames of the task \textit{Put the fruits into the basket}.]{
\includegraphics[ width=0.9\linewidth]{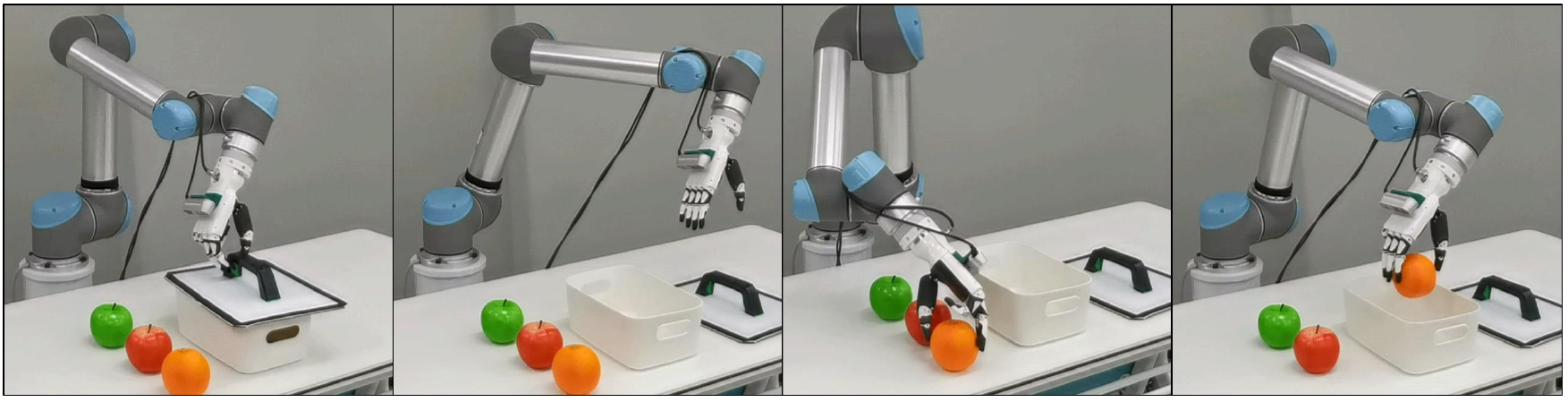}
\label{subfig:longHorizonDemo1}
}
\subfigure[Key frames of the task \textit{Open the drawer and pick out the objects inside}.]{
\includegraphics[width=0.9\linewidth]{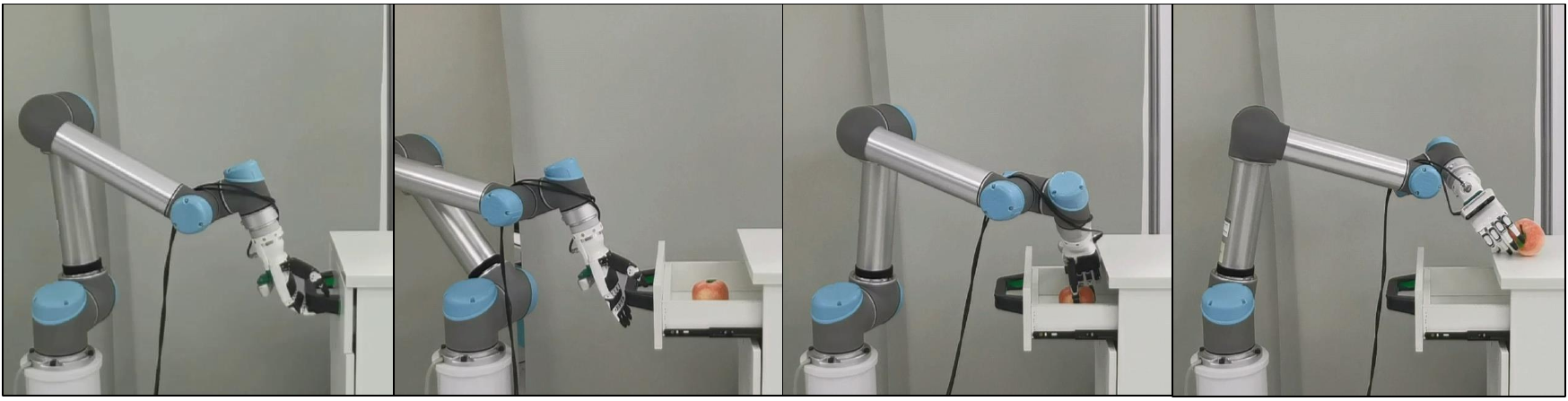}
\label{subfig:longHorizonDemo2}
}
\caption{Demonstration of long-horizon dexterous manipulation. The input of the RoboDexVLM framework is one sentence describing the task to be completed. The relevant skills are invoked automatically to interact with the objects for the open-vocabulary task. The corresponding videos are accessible in our \MYhref[blue]{https://henryhcliu.github.io/RoboDexVLM}{project page}.}
\label{fig:long-horizonPlanning}
\end{figure*}

Moreover, Fig.~\ref{fig:long-horizonPlanning} reveals the key frames of two long-horizon manipulation tasks. In Fig.~\ref{subfig:longHorizonDemo1}, the robot is instructed to place all the fruits into a lidded basket. The VLM generates the sequence of actions, which involves setting the lid aside before placing all the fruits into the open basket, following a logical reasoning process tailored for the open-vocabulary task. Another challenging manipulation task \textit{Open the drawer and pick out the objects inside} is illustrated in Fig.~\ref{subfig:longHorizonDemo2}. The robot first detects and grasps the handle of the drawer before pulling it open to identify the objects stored inside. Once the objects are detected, the robot manipulator adopts a gentle dexterous hand posture to retrieve them and place them onto the cabinet surface.

In summary, the RoboDexVLM manipulation system shows particular adeptness in handling slim or oddly shaped objects, showcasing its advanced dexterity and adaptive operating mechanisms. 
The demonstration of dexterous manipulation in long-horizon tasks offers valuable insights into human-like grasping and the sequences involved in task execution.
\subsection{Comparison and Ablation Study}

\begin{table}[t]
  \centering
  \caption{Comparison on Object Detection with Diverse Labels with an Adjective for Open-Vocabulary Manipulation.}
    \begin{tabular}{ccccc}
    \toprule
    \multicolumn{1}{c}{\textbf{Method}} & \multicolumn{1}{l}{\textbf{Label}} & \multicolumn{1}{l}{\textbf{Succ. Rate (\%)}} & \multicolumn{1}{l}{\textbf{Inf. Time (ms)} } \\
    \midrule
    \multirow{3}{*}{with YOLO v11}
          & $1$ & $53.33$& $10.7\pm0.2$\\
          & $2$ & $46.67$ & $10.6\pm0.2$ \\
          & $3$ & $46.67$ & $10.8\pm0.3$\\
    \midrule
    \multirow{3}{*}{RoboDexVLM (Ours)} 
          & $1$ & $100.00$ & $372.6\pm10.5$ \\
          & $2$ & $93.33$ &$388.2\pm11.4$\\
          & $3$ & $100.00$ & $392.1\pm11.7$ \\
    \bottomrule
    \end{tabular}%
  \label{tab:AblationYOLO}%
\begin{flushleft}
    {\small \textbf{Remark:} \textit{Label 1}:\{red apple\}; \textit{Label 2}: \{the middle carambola\}; \textit{Label 3}: \{the smaller carambola\}.}
\end{flushleft}
\end{table}%
\begin{table}[t]
  \centering
  \caption{Effectiveness Evaluation of Recovery Mechanism.}
    \begin{tabular}{ccccc}
    \toprule
    \multicolumn{1}{c}{\textbf{Method}} & \multicolumn{1}{l}{\textbf{Task Category}} & \multicolumn{1}{l}{\textbf{Succ. Rate (\%)}} & \multicolumn{1}{l}{\textbf{Exec. Time (s)} } \\
    \midrule
    \multirow{3}{*}{w/o Recovery}
          & $1$ & $90.00$& $30.5\pm2.6$\\
          & $2$ & $20.00$ & $ 31.1\pm3.0$ \\
          & $3$ & $66.67$ & $121.4\pm5.8$\\
    \midrule
    \multirow{3}{*}{w/ Recovery} 
          & $1$ & $96.67$ & $31.5\pm2.8$ \\
          & $2$ & $96.67$ &$ 32.7\pm3.9$\\
          & $3$ & $93.33$ & $129.4\pm6.2$ \\
    \bottomrule
    \end{tabular}%
  \label{tab:RecEffect}%
\begin{flushleft}
    {\small \textbf{Remark:} \textit{Task Category 1}: single fruit semantic sorting (grasping failure); \textit{Task Category 2}: single fruit semantic sorting (object position changed); \textit{Task Category 3}: multi-object arrangement.}
\end{flushleft}
\end{table}%

\begin{table*}[t]
  \centering
  \caption{Robustness and Stability Evaluation of RoboDexVLM with Diverse Open-Vocabulary Tasks.}
    \begin{tabular}{cccccc}
    \toprule
    \multirow{2}{*}{\textbf{Task Description}} 
          & \textbf{Skill Seq.} & \textbf{Succ. Rate} &  \textbf{Succ. Rate} & \multirow{2}{*}{\textbf{Reasoning Time (s)}} & \multirow{2}{*}{\textbf{Exec. Time (s)}}\\
          & \textbf{Length} & \textbf{w/o Memory (\%)} & \textbf{w/ Memory (\%)} & & \\
    \midrule
    "Put the green apple in the basket."& $8$ & $70.00$& $95.00$ &\multirow{3}{*}{$18.2\pm5.1$}&\multirow{3}{*}{$31.5\pm2.8$}\\
    "Put carambola in the middle in the box." & $8$ & $65.00$ & $90.00$ & & \\
    "Put the smaller carambola in the bowl." & $8$ & $75.00$ & $95.00$ & & \\
    \midrule
    "Place the bowl in the drawer." & $14$ & $40.00$& $90.00$ &\multirow{2}{*}{$28.5\pm7.3$}&\multirow{2}{*}{$63.6\pm3.3$}\\
    "Put the peach in the drawer on the table top."& $14$ & $35.00$ & $85.00$ & & \\
    \midrule
    "Put all the fruits in the basket." (without a lid)& $24$ & $25.00$& $85.00$ &\multirow{2}{*}{$35.7\pm9.9$}&{{$98.4\pm4.9$}}\\
    "Put all the fruits in the box." (with a lid) & $30$ & $20.00$ & $85.00$ & & $121.4\pm5.4$\\
    \bottomrule
    \end{tabular}%
  \label{tab:RobustStabilityAnalysis}%
\end{table*}%
To evaluate the superiority of using a language-grounded open-world object segmentation module over traditional predefined and trained object detection approaches for the foundation abilities catered to open-vocabulary tasks. 
Specifically, this study examines the contribution of the open-world object segmentation module enabled by LangSAM~\cite{luca-medeiros_lang-segment-anything} within the RoboDexVLM framework, comparing its performance against the classical object detection approach, YOLOv11~\cite{make5040083}. The evaluation encompasses a range of diverse open-world object detection tasks characterized by varying color, spatial, and size attributes.
As shown in Table~\ref{tab:AblationYOLO}, our language-driven method demonstrates an improved success rate across all evaluated task scenarios under identical environmental conditions.  
Although with significantly faster inference time, the results reveal severe limitations of the YOLO detection pipelines when confronted with nuanced manipulation requirements inherent to open-vocabulary settings. While both approaches share comparable computational hardware constraints, our method achieved near-perfect success rates versus baseline scores below 55\%. This discrepancy stems from two critical factors, semantic grounding fidelity and attribute reasoning.
First, language-conditioned queries (e.g., “smaller carambola”) enable pixel-precise localization where conventional bounding boxes fail.  
Second, explicit modeling of relational descriptors eliminates heuristic positional filtering required by static-class detectors, reducing cascading errors during multi-object interactions.  
These findings conclusively validate that transitioning from rigid taxonomy
to language-anchored segmentation paradigms fundamentally elevates robotic systems’ operational envelopes, a prerequisite for deploying general-purpose manipulators in unstructured environments.

In addition, we evaluate the effectiveness of the recovery mechanism through performance metrics and comparisons with a non-recoverable approach, highlighting its impact on the reliability of robot tasks, as shown in Table~\ref{tab:RecEffect}.
The results indicate that the recovery mechanism significantly enhances task performance. For task 1, single fruit semantic sorting (grasping failure), RoboDexVLM with recovery achieved a success rate of 96.67\%, compared to 90.00\% without recovery. This marginal improvement in success rate, coupled with a comparable execution time of $31.5 \pm 2.4\,\text{s}$ versus $30.5 \pm 2.3\,\text{s}$, suggests that even relatively simple tasks benefit from the error-correcting capabilities of the recovery mechanism, allowing for slight adjustments that improve overall reliability.
In contrast, task 2, single fruit semantic sorting (object
position changed), demonstrates a more pronounced impact of the recovery mechanism. With comparable execution time, the success rate increased from 20.00\% without recovery to 96.67\% with recovery, reflecting a substantial improvement in task execution under disturbing conditions.  
For the last category of tasks,  multiple object arrangement, the success rate with the recovery mechanism is higher than the non-recovery approach by 26.66\%, which implies that for long-sequence and complex tasks, reflection and replan from failures is especially necessary.
The results from all tasks clearly illustrate that while the recovery mechanism may incur additional execution time in more complex scenarios, it ultimately enhances the reliability and success of the robot’s actions. This trade-off between efficiency and effectiveness underscores the importance of incorporating recovery strategies, particularly for tasks requiring higher degrees of precision and adaptability. 
Moreover, the effectiveness of the memory storage is further evaluated in Table~\ref{tab:RobustStabilityAnalysis} across diverse of open-vocabulary task descriptions. 
For simple atomic tasks, the average success rate without memory is approximately 25.00\% lower than that of the version utilizing memory. As task complexity and the length of skill sequences increase, the advantages of the memory module become increasingly evident. Notably, when addressing tasks that involve lid opening operations, the version without memory retrieval achieves a success rate of only 20.00\%, whereas the full version of RoboDexVLM attains a success rate of 85.00\%.

\subsection{Robustness and Hazard Analysis of RoboDexVLM}
To evaluate the robustness of the RoboDexVLM framework, we conduct quantitative experiments on three categories of open-vocabulary tasks spanning varying complexity levels as shown in Table \ref{tab:RobustStabilityAnalysis}. The results reveal critical relationships between task structure, reasoning efficiency, and physical execution reliability.

The single fruit semantic sorting tasks in the first three rows of Table~\ref{tab:RobustStabilityAnalysis} demonstrated superior reliability with more than 90.00\% success rate over 30 trials, supported by efficient reasoning and execution. This aligns with our hypothesis that atomic object manipulations minimize cumulative uncertainty.
In contrast, tasks related to object retrieval from drawers and putting into drawers in the fourth and fifth row of Table~\ref{tab:RobustStabilityAnalysis} exhibited slightly reduced success rates, attributable to compounded challenges. The extended reasoning time ($28.5\pm7.3$\,s) versus the first kind of task further reflects a more comprehensive and longer reasoning process for task planning.
Ultimately, while the third category of tasks, long-horizon multi-object arrangement, demanded substantially longer reasoning time, its success rate is comparable with drawer operations due to easier grasping of the vertical handle of a basket compared with the horizontal handle with less safety margin.
Lastly, the reasoning time for an order of skill $\mathcal{O}_\tau$ gradually increases with higher task complexity, as shown from top to down in Table~\ref{tab:RobustStabilityAnalysis}. This observation is consistent with the progressively increasing reasoning time and execution time.

\section{Conclusion}
This paper presents RoboDexVLM, a novel framework for dexterous manipulation that integrates dynamically updated variable storage mechanisms with interaction primitives to address the challenges of open-vocabulary and long-horizon tasks. By unifying VLMs with modular skill libraries and test-time adaptation, our system demonstrates robust adaptability across diverse scenarios, from atomic object manipulations to complex multi-stage operations. 
Key innovations include a hierarchical recovery mechanism that mitigates cascading errors in complex or long-horizon tasks, language-anchored segmentation paradigms enabling precise attribute-based object dexterous grasp perception, and closed-loop perception-action pipelines optimized for real-world dexterous manipulation deployment. 
By decoupling task planning from low-level control through reusable primitives, RoboDexVLM lowers the barrier for non-expert users to program complex dexterous robotic behaviors via natural language commands. Future work will focus on scaling this paradigm to dynamic multi-agent collaboration scenarios and enhancing failure prediction using causal reasoning models. This research marks a pivotal step toward general-purpose manipulation systems capable of operating reliably with minimal reconfiguration efforts.
\bibliographystyle{IEEEtran}
\bibliography{refs}

\begin{thebibliography}{10}
\providecommand{\url}[1]{#1}
\csname url@samestyle\endcsname
\providecommand{\newblock}{\relax}
\providecommand{\bibinfo}[2]{#2}
\providecommand{\BIBentrySTDinterwordspacing}{\spaceskip=0pt\relax}
\providecommand{\BIBentryALTinterwordstretchfactor}{4}
\providecommand{\BIBentryALTinterwordspacing}{\spaceskip=\fontdimen2\font plus
\BIBentryALTinterwordstretchfactor\fontdimen3\font minus \fontdimen4\font\relax}
\providecommand{\BIBforeignlanguage}[2]{{%
\expandafter\ifx\csname l@#1\endcsname\relax
\typeout{** WARNING: IEEEtran.bst: No hyphenation pattern has been}%
\typeout{** loaded for the language `#1'. Using the pattern for}%
\typeout{** the default language instead.}%
\else
\language=\csname l@#1\endcsname
\fi
#2}}
\providecommand{\BIBdecl}{\relax}
\BIBdecl

\bibitem{make5040083}
J.~Terven, D.-M. Córdova-Esparza, and J.-A. Romero-González, ``A comprehensive review of {YOLO} architectures in computer vision: From {YOLOv1} to {YOLOv8} and {YOLO-NAS},'' \emph{Machine Learning and Knowledge Extraction}, vol.~5, no.~4, pp. 1680--1716, 2023.

\bibitem{liu2024grounding}
S.~Liu, Z.~Zeng, T.~Ren, F.~Li, H.~Zhang, J.~Yang, Q.~Jiang, C.~Li, J.~Yang, H.~Su \emph{et~al.}, ``Grounding {DINO}: Marrying {DINO} with grounded pre-training for open-set object detection,'' in \emph{European Conference on Computer Vision}, 2024, pp. 38--55.

\bibitem{kirillov2023segment}
A.~Kirillov, E.~Mintun, N.~Ravi, H.~Mao, C.~Rolland, L.~Gustafson, T.~Xiao, S.~Whitehead, A.~C. Berg, W.-Y. Lo \emph{et~al.}, ``Segment anything,'' in \emph{Proceedings of the IEEE/CVF International Conference on Computer Vision}, 2023, pp. 4015--4026.

\bibitem{luca-medeiros_lang-segment-anything}
L.~Medeiros, ``Language segment-anything: {SAM} with text prompt,'' \url{https://github.com/luca-medeiros/lang-segment-anything}, 2024.

\bibitem{jin2024robotgpt}
Y.~Jin, D.~Li, A.~Yong, J.~Shi, P.~Hao, F.~Sun, J.~Zhang, and B.~Fang, ``Robot{GPT}: Robot manipulation learning from {ChatGPT},'' \emph{IEEE Robotics and Automation Letters}, vol.~9, no.~3, pp. 2543--2550, 2024.

\bibitem{gao2024physically}
J.~Gao, B.~Sarkar, F.~Xia, T.~Xiao, J.~Wu, B.~Ichter, A.~Majumdar, and D.~Sadigh, ``Physically grounded vision-language models for robotic manipulation,'' in \emph{2024 IEEE International Conference on Robotics and Automation}, 2024, pp. 12\,462--12\,469.

\bibitem{firoozi2023foundation}
R.~Firoozi, J.~Tucker, S.~Tian, A.~Majumdar, J.~Sun, W.~Liu, Y.~Zhu, S.~Song, A.~Kapoor, K.~Hausman \emph{et~al.}, ``Foundation models in robotics: Applications, challenges, and the future,'' \emph{The International Journal of Robotics Research}, pp. 1--33, 2024.

\bibitem{huang2024rekep}
W.~Huang, C.~Wang, Y.~Li, R.~Zhang, and L.~Fei-Fei, ``{ReKep}: Spatio-temporal reasoning of relational keypoint constraints for robotic manipulation,'' \emph{arXiv preprint arXiv:2409.01652}, 2024.

\bibitem{pan2025omnimanip}
M.~Pan, J.~Zhang, T.~Wu, Y.~Zhao, W.~Gao, and H.~Dong, ``{OmniManip}: Towards general robotic manipulation via object-centric interaction primitives as spatial constraints,'' \emph{arXiv preprint arXiv:2501.03841}, 2025.

\bibitem{liu2024robomamba}
J.~Liu, M.~Liu, Z.~Wang, P.~An, X.~Li, K.~Zhou, S.~Yang, R.~Zhang, Y.~Guo, and S.~Zhang, ``{RoboMamba}: Efficient vision-language-action model for robotic reasoning and manipulation,'' in \emph{The 38th Annual Conference on Neural Information Processing Systems}, 2024.

\bibitem{shridhar2022cliport}
M.~Shridhar, L.~Manuelli, and D.~Fox, ``{CLIP}ort: What and where pathways for robotic manipulation,'' in \emph{Conference on Robot Learning}, 2022, pp. 894--906.

\bibitem{pmlr-v229-huang23b}
W.~Huang, C.~Wang, R.~Zhang, Y.~Li, J.~Wu, and L.~Fei-Fei, ``Vox{P}oser: Composable {3D} value maps for robotic manipulation with language models,'' in \emph{The 7th Conference on Robot Learning}, vol. 229, 2023, pp. 540--562.

\bibitem{ha2023scaling}
H.~Ha, P.~Florence, and S.~Song, ``Scaling up and distilling down: Language-guided robot skill acquisition,'' in \emph{Conference on Robot Learning}, 2023, pp. 3766--3777.

\bibitem{liu2008multisensory}
H.~Liu, K.~Wu, P.~Meusel, N.~Seitz, G.~Hirzinger, M.~Jin, Y.~Liu, S.~Fan, T.~Lan, and Z.~Chen, ``Multisensory five-finger dexterous hand: The {DLR/HIT} hand {II},'' in \emph{2008 IEEE/RSJ International Conference on Intelligent Robots and Systems}, 2008, pp. 3692--3697.

\bibitem{kim2021integrated}
U.~Kim, D.~Jung, H.~Jeong, J.~Park, H.-M. Jung, J.~Cheong, H.~R. Choi, H.~Do, and C.~Park, ``Integrated linkage-driven dexterous anthropomorphic robotic hand,'' \emph{Nature Communications}, vol.~12, no.~1, p. 7177, 2021.

\bibitem{liu2021research}
H.~Liu, Z.~Liu, H.~Liu, and W.~Lin, ``Research on robot visual grabbing based on mechanism analysis,'' in \emph{2021 IEEE 11th Annual International Conference on Cyber Technology in Automation, Control, and Intelligent Systems}, 2021, pp. 181--186.

\bibitem{dogar2010push}
M.~R. Dogar and S.~S. Srinivasa, ``Push-grasping with dexterous hands: Mechanics and a method,'' in \emph{2010 IEEE/RSJ International Conference on Intelligent Robots and Systems}, 2010, pp. 2123--2130.

\bibitem{morrison2020learning}
D.~Morrison, P.~Corke, and J.~Leitner, ``Learning robust, real-time, reactive robotic grasping,'' \emph{The International Journal of Robotics Research}, vol.~39, no. 2-3, pp. 183--201, 2020.

\bibitem{sundermeyer2021contact}
M.~Sundermeyer, A.~Mousavian, R.~Triebel, and D.~Fox, ``{Contact-GraspNet}: Efficient 6-{DoF} grasp generation in cluttered scenes,'' in \emph{2021 IEEE International Conference on Robotics and Automation}, 2021, pp. 13\,438--13\,444.

\bibitem{fang2023anygrasp}
H.-S. Fang, C.~Wang, H.~Fang, M.~Gou, J.~Liu, H.~Yan, W.~Liu, Y.~Xie, and C.~Lu, ``{AnyGrasp}: Robust and efficient grasp perception in spatial and temporal domains,'' \emph{IEEE Transactions on Robotics}, vol.~39, no.~5, pp. 3929--3945, 2023.

\bibitem{wei2024mathcaldro}
Z.~Wei, Z.~Xu, J.~Guo, Y.~Hou, C.~Gao, C.~Zhehao, J.~Luo, and L.~Shao, ``{D(R,O) Grasp}: A unified representation of robot and object interaction for cross-embodiment dexterous grasping,'' in \emph{CoRL Workshop on Learning Robot Fine and Dexterous Manipulation: Perception and Control}, 2024.

\bibitem{chen2022learning}
Q.~Chen, K.~V. Wyk, Y.-W. Chao, W.~Yang, A.~Mousavian, A.~Gupta, and D.~Fox, ``Learning robust real-world dexterous grasping policies via implicit shape augmentation,'' in \emph{6th Annual Conference on Robot Learning}, 2022.

\bibitem{han2024learning}
Y.~Han, Z.~Chen, K.~A. Williams, and H.~Ravichandar, ``Learning prehensile dexterity by imitating and emulating state-only observations,'' \emph{IEEE Robotics and Automation Letters}, vol.~9, no.~10, pp. 8266 -- 8273, 2024.

\bibitem{wang2024dexcap}
C.~Wang, H.~Shi, W.~Wang, R.~Zhang, L.~Fei-Fei, and C.~K. Liu, ``Dexcap: Scalable and portable {Mocap} data collection ystem for dexterous manipulation,'' \emph{arXiv preprint arXiv:2403.07788}, 2024.

\bibitem{chen2023bi}
Y.~Chen, Y.~Geng, F.~Zhong, J.~Ji, J.~Jiang, Z.~Lu, H.~Dong, and Y.~Yang, ``Bi-dexhands: Towards human-level bimanual dexterous manipulation,'' \emph{IEEE Transactions on Pattern Analysis and Machine Intelligence}, vol.~46, no.~5, pp. 2804--2818, 2023.

\bibitem{ma2024survey}
Y.~Ma, Z.~Song, Y.~Zhuang, J.~Hao, and I.~King, ``A survey on vision-language-action models for embodied {AI},'' \emph{arXiv preprint arXiv:2405.14093}, 2024.

\bibitem{ak2023learning}
A.~C. Ak, E.~E. Aksoy, and S.~Sariel, ``Learning failure prevention skills for safe robot manipulation,'' \emph{IEEE Robotics and Automation Letters}, vol.~8, no.~12, pp. 7994--8001, 2023.

\bibitem{xiong2024aic}
C.~Xiong, C.~Shen, X.~Li, K.~Zhou, J.~Liu, R.~Wang, and H.~Dong, ``{AIC MLLM}: Autonomous interactive correction {MLLM} for robust robotic manipulation,'' \emph{arXiv preprint arXiv:2406.11548}, 2024.

\bibitem{ravi2025sam}
N.~Ravi, V.~Gabeur, Y.-T. Hu, R.~Hu, C.~Ryali, T.~Ma, H.~Khedr, R.~R{\"a}dle, C.~Rolland, L.~Gustafson, E.~Mintun, J.~Pan, K.~V. Alwala, N.~Carion, C.-Y. Wu, R.~Girshick, P.~Dollar, and C.~Feichtenhofer, ``{SAM} 2: Segment anything in images and videos,'' in \emph{The 13th International Conference on Learning Representations}, 2025.

\bibitem{hurst2024gpt}
A.~Hurst, A.~Lerer, A.~P. Goucher, A.~Perelman, A.~Ramesh, A.~Clark, A.~Ostrow, A.~Welihinda, A.~Hayes, A.~Radford \emph{et~al.}, ``{GPT-4o} system card,'' \emph{arXiv preprint arXiv:2410.21276}, 2024.

\end{thebibliography}

\end{document}